\documentclass[sigconf]{acmart}
\usepackage{makecell}
\usepackage{multirow}
\usepackage{wrapfig}
\usepackage{graphicx}
\usepackage{breakurl}
\usepackage{ifthen}
\usepackage{xspace}
\usepackage[linesnumbered,ruled,vlined]{algorithm2e}
\usepackage{algpseudocode}
\SetKwInput{Input}{Input}
\SetKwProg{kwServer}{ServerUpdate}{:}{}
\SetKwProg{kwClient}{ClientUpdate}{:}{}

\usepackage{pifont}
\newcommand{\ours}{\texttt{FedHelp}\xspace}
\usepackage{colortbl}
\AtBeginDocument{%
  }

\copyrightyear{2025}
\acmYear{2025}
\setcopyright{acmlicensed}\acmConference[KDD '25]{Proceedings of the 31st ACM SIGKDD Conference on Knowledge Discovery and Data Mining V.1}{August 3--7, 2025}{Toronto, ON, Canada}
\acmBooktitle{Proceedings of the 31st ACM SIGKDD Conference on Knowledge Discovery and Data Mining V.1 (KDD '25), August 3--7, 2025, Toronto, ON, Canada}
\acmDOI{10.1145/3690624.3709235}
\acmISBN{979-8-4007-1245-6/25/08}

\begin{document}


\title{Asymmetrical Reciprocity-based Federated Learning for Resolving Disparities in Medical Diagnosis}








\author{Jiaqi Wang*}
\affiliation{%
  \institution{The Pennsylvania State University}
  \city{University Park}
  \state{PA}
    \country{USA}}
\email{jqwang@psu.edu}
\author{Ziyi Yin*}
\affiliation{%
  \institution{The Pennsylvania State University}
    \city{University Park}
  \state{PA}
  \country{USA}}
\email{zmy5171@psu.edu}
\thanks{*These authors contributed equally to this work.}
\author{Quanzeng You}
\affiliation{%
  \institution{ByteDance}
  \city{Bellevue}
  \state{WA}
  \country{USA}}
\email{quanzeng.you@outlook.com}

\author{Lingjuan Lyu}
\affiliation{%
  \institution{Sony AI}
  \city{Zurich}
  \country{Switerzland}}
\email{Lingjuan.Lv@sony.com}
\author{Fenglong Ma}
\affiliation{%
  \institution{The Pennsylvania State University}
    \city{University Park}
  \state{PA}
  \country{USA}}
\email{fenglong@psu.edu}

\renewcommand{\shortauthors}{Wang et al.}

\begin{abstract}
   Geographic health disparities pose a pressing global challenge, particularly in underserved regions of low- and middle-income nations. Addressing this issue requires a collaborative approach to enhance healthcare quality, leveraging support from medically more developed areas. Federated learning emerges as a promising tool for this purpose. However, the scarcity of medical data and limited computation resources in underserved regions make collaborative training of powerful machine learning models challenging. Furthermore, there exists an asymmetrical reciprocity between underserved and developed regions. To overcome these challenges, we propose a novel cross-silo federated learning framework, named \ours, aimed at alleviating geographic health disparities and fortifying the diagnostic capabilities of underserved regions. Specifically, \ours leverages foundational model knowledge via one-time API access to guide the learning process of underserved small clients, addressing the challenge of insufficient data. Additionally, we introduce a novel asymmetric dual knowledge distillation module to manage the issue of asymmetric reciprocity, facilitating the exchange of necessary knowledge between developed large clients and underserved small clients. We validate the effectiveness and utility of \ours through extensive experiments on both medical image classification and segmentation tasks. The experimental results demonstrate significant performance improvement compared to state-of-the-art baselines, particularly benefiting clients in underserved regions.
\end{abstract}

%
%




\maketitle

\section{Introduction}
Geographic health disparities pose a fundamental challenge to countries worldwide~\cite{ruger2006global,mackinnon2023mapping,weisent2012socioeconomic,doogan2018validation}. These disparities underscore the unequal distribution of health resources, access to healthcare services, and health outcomes across different geographic regions. Particularly in low- and middle-income nations, rural areas often grapple with significant challenges related to healthcare infrastructure, access to medical professionals, and essential health services~\cite{wang2012geographical,ayuningtyas2022geographic}. This lack of access can result in higher rates of preventable diseases, maternal and child mortality, and overall poorer health outcomes in these regions. Furthermore, factors such as limited funding, lack of education, and a shortage of expertise in rural areas hinder their ability to invest in cutting-edge technologies. Consequently, \textit{finding collaborative ways to enhance healthcare quality in underserved regions with the support of medically developed areas is an urgent and essential social issue}.



Federated learning, a technique widely employed in the medical domain, presents a potential solution to this challenge by enabling collaborative training of robust machine learning models without centralizing healthcare data~\cite{wang2022towards,xu2022closing, wang2023federated,liu2021feddg,jiang2023fair,mcmahan2017communication,mendieta2022local,li2021model,tang2022fedcor}. However, many existing approaches necessitate clients to employ identical models, a requirement unsuitable for our context where underserved regions face economic constraints in procuring high-cost computational resources. In essence, these regions or clients can only afford small-sized models. In contrast, medically developed areas typically utilize large models, thereby resulting in the challenge of \textit{heterogeneous models} in federated learning. 
While numerous strategies~\cite{wang2023towards,ilhan2023scalefl,huang2022learn,ma2022layer,li2019fedmd,yi2023fedgh,yu2022resource,liu2022no} have been proposed to tackle the challenges posed by heterogeneous federated learning, these approaches still suffer from the following challenges:

\begin{figure}[t]
  \centering
  \includegraphics[width=0.48\textwidth]{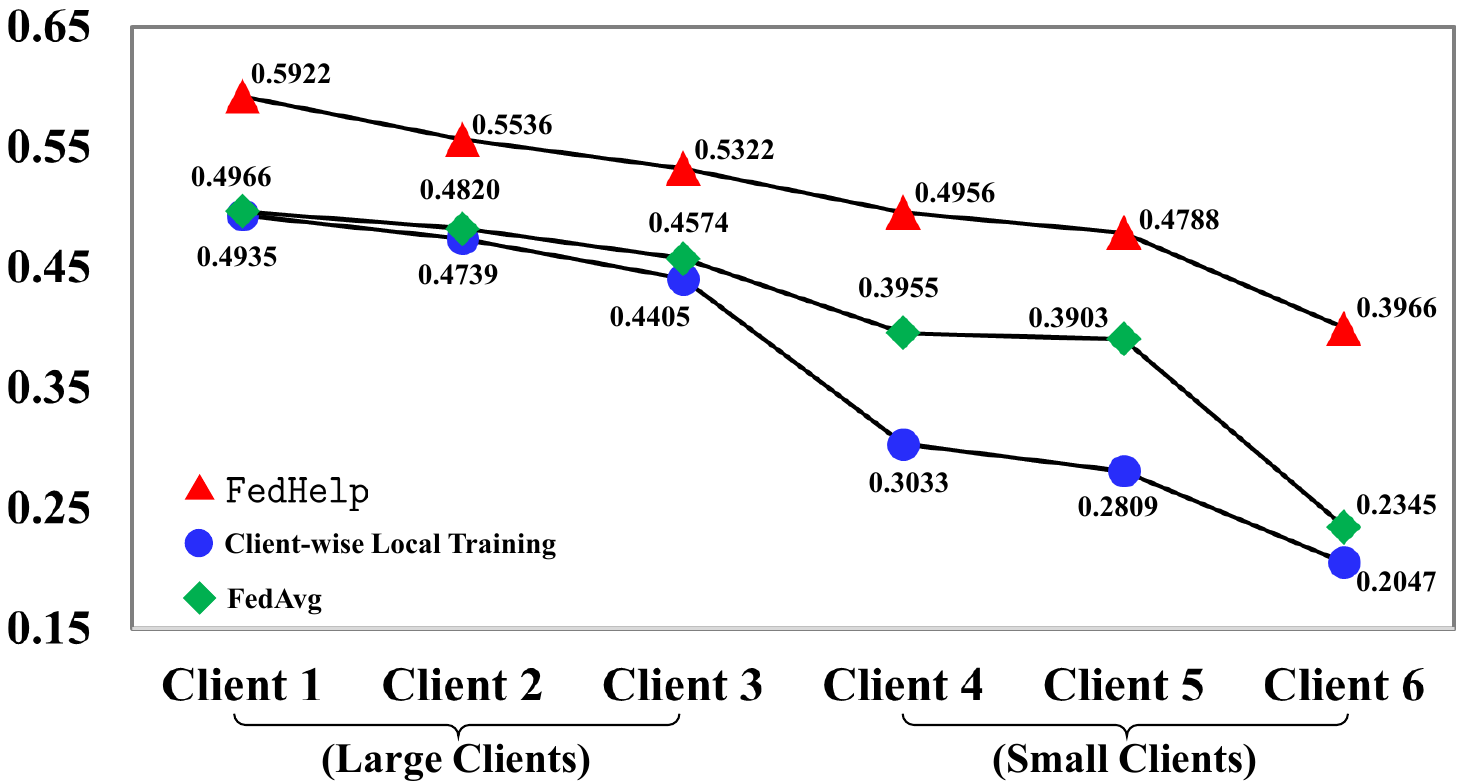}
  \caption{Client accuracy comparison between client-wise local training, FedAvg, and the proposed \ours on the Fed-ISIC19 dataset. The size of each client can be found in Section~\ref{sec:melanoma}.}
  \vspace{-0.15in}
  \label{fig:motivation}
\end{figure}

\begin{figure*}[t]
\centering
\includegraphics[width=0.95\textwidth]{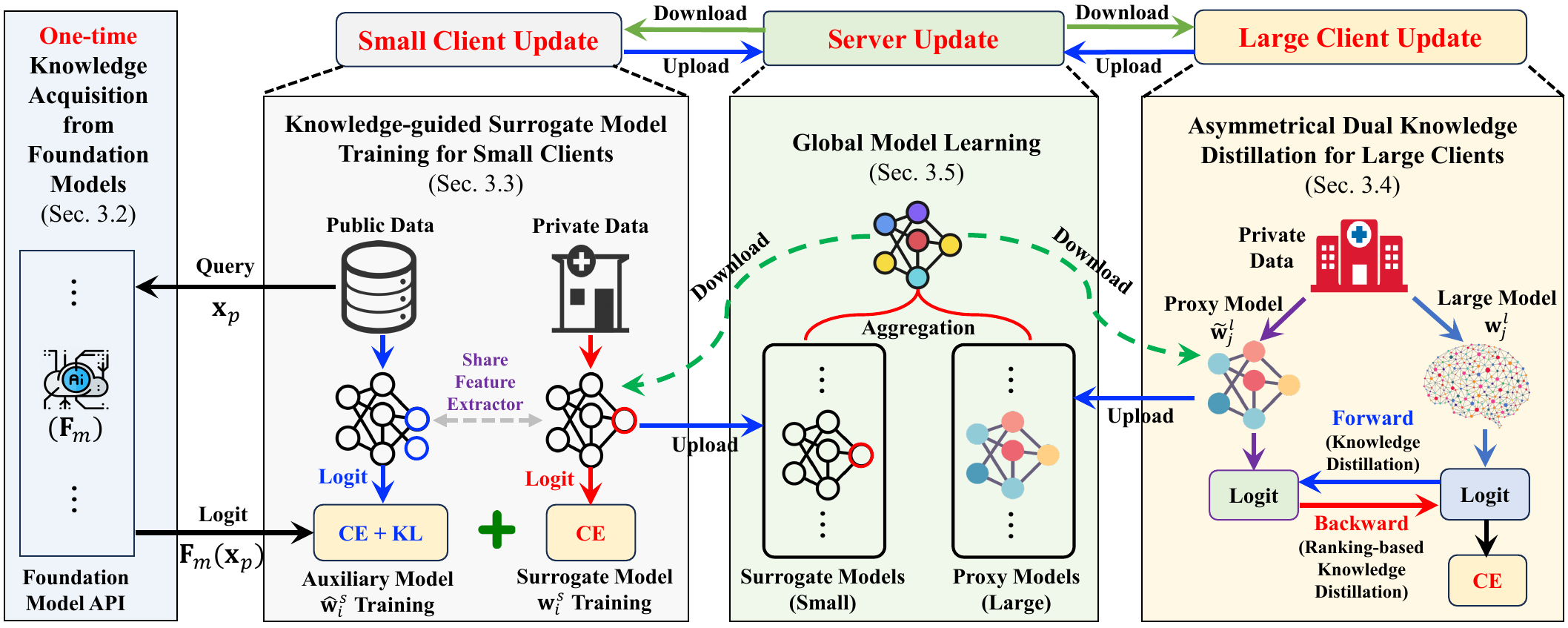}
    \caption{Overview of the proposed \ours framework. ``CE''/``KL'' denotes the cross-entropy loss/Kullback–Leibler divergence.}
    \label{fig:overview}
\end{figure*}

$\bullet$ \emph{\textbf{C1}: Limited medical data in underserved regions.}
Several factors, including inadequate access to healthcare facilities, a lack of awareness regarding the importance of medical record-keeping, and insufficient technology for maintaining electronic health records, contribute to the scarcity of medical data in underserved regions. In our federated learning framework, each region is treated as a client. As depicted in Figure~\ref{fig:motivation}, small clients typically exhibit significantly lower performance than larger ones in local training. While applying the existing federated learning algorithm FedAvg~\cite{mcmahan2017communication} does improve performance, the benefits are primarily observed in clients with relatively small datasets (clients 4 and 5), excluding both the largest and smallest ones. Consequently, it is crucial to explore new learning strategies that can benefit \textbf{ALL} clients in our setting. Without such inclusive approaches, there is little incentive for medically developed large clients to contribute to the improvement of underserved small clients.



$\bullet$ \emph{\textbf{C2}: Asymmetrical reciprocity among clients.}
Traditional federated learning methodologies treat all clients equally, aiming to collaboratively train a shared global model or personalized client models. However, this study deviates from the norm by focusing on harnessing the abundant resources of medically developed areas to enhance the diagnostic performance of underserved regions using compact models without the need to share their data. In essence, the small clients emerge as the primary beneficiaries. While large clients can gain insights from small client information through model aggregation, they still act as valuable resource providers. 
Thus, the collaboration in this scenario exhibits asymmetric reciprocity among clients, which aligns with our observations in Figure~\ref{fig:motivation}. However, effectively modeling this asymmetric reciprocity poses a novel challenge in federated learning.

To tackle these challenges simultaneously, we introduce a groundbreaking cross-silo \underline{\textbf{fed}}erated learning framework called \ours\footnote{The source code is available at \url{https://github.com/JackqqWang/fedhelp}.}, specifically tailored to combat geographic \underline{\textbf{he}}a\underline{\textbf{l}}th dis\underline{\textbf{p}}arities and bolster the diagnostic capabilities of underserved regions, as illustrated in Figure~\ref{fig:overview}. 
To tackle the first challenge (\textbf{C1}) encountered during the training of small clients, we advocate harnessing knowledge from foundational models~\cite{zhuang2023foundation} via only one-time API access using public data rather than private medical data in Section~\ref{sec:knowledge_acq}. This acquired knowledge serves as a guiding light in training small surrogate models in Section~\ref{sec:small_training}. Meanwhile, the second challenge (\textbf{C2}) arises when training large clients. To circumvent escalating communication costs, we advocate distilling a proxy model for each large client, mirroring the network structure of small clients. This proxy model acts as an intermediary, facilitating the exchange of information between large and small clients. Specifically, we have devised an innovative asymmetrical dual knowledge distillation strategy to address the challenge of asymmetrical reciprocity in Section~\ref{sec:large_training}. Subsequently, the small surrogate and proxy models are uploaded to the server for aggregation in Section~\ref{sec:global_model_aggregation}. 

To the best of our knowledge, this is the first work to leverage federated learning techniques to mitigate the global issue of geographic health disparities, thereby augmenting healthcare quality in underserved regions. In particular, we introduce a unique framework, \ours, designed to address the challenge of small data through accessing expansive foundation models via API in an efficient way and tackle the distinctive hurdle of asymmetrical reciprocity via the proposed dual knowledge distillation strategy.
We conduct comprehensive experiments encompassing multi-class and binary medical image classification tasks as well as 2D and 3D semantic segmentation tasks, comparing the results with state-of-the-art baselines. The experimental outcomes unequivocally affirm the efficacy of the \ours framework.

\section{Related Work}
\subsection{Federated Learning}
Federated learning~\cite{mcmahan2017communication,che2023multimodal,zhou2022adversarial}, aiming to collaboratively train a machine learning model without sharing clients' private data, has been applied to the medical domain~\cite{wang2022towards,xu2022closing, wang2023federated,liu2021feddg,jiang2023fair,wang2023federated,wang2024fedmeki,wang2024fedkim}. Compared with traditional federated learning frameworks~\cite{mcmahan2017communication,mendieta2022local,li2021model,tang2022fedcor}, personalized federated learning focuses on the performance of the local clients~\cite{fallah2020personalized,t2020personalized,zhang2022personalized,shen2022cd2}, achieving superior performance.
However, either traditional federated learning or personalized federated learning models require that all the clients share the identical model structure. Though several heterogeneous federated learning frameworks~\cite{ilhan2023scalefl,huang2022learn,ma2022layer,li2019fedmd,yi2023fedgh,yu2022resource,wang2024towards, wangpfedclub} have been proposed to solve this issue, they are not typically designed to address the challenges in the medical domain. While one recent medical-related work ProxyFL~\cite{kalra2023decentralized} proposes a decentralized federated learning method with each client managing a small model, it neglects the essential distinctions between private and proxy models. Moreover, this decentralized approach leads to increased communication expenses.

\subsection{Dual Knowledge Distillation}
Knowledge distillation~\cite{hinton2015distilling} treats the large model as a teacher, which passes knowledge to a small student model to enhance its performance. The most relevant work is bidirectional or dual knowledge distillation~\cite{kweon2021bidirectional}, enabling the teacher and student to learn knowledge from each other. In~\cite{reddi2021rankdistil, kweon2021bidirectional}, the bidirectional distillation technique is utilized to solve the top-k ranking research problem and machine translation~\cite{inaguma2021source,zhang2021dual,zhuang-tu-2023-pretrained}. Although a few studies apply bidirectional knowledge distillation in federated learning to conduct the tasks of distracted driving detection~\cite{shang2023fedbikd}, medical relation extraction~\cite{sui2020feded}, and the IoT system~\cite{qi2022fedbkd}, they all treat both teacher and student models equally yet ignore the importance of asymmetrical reciprocity.

FedType~\cite{pmlr-v235-wang24cs} is the most relevant work, addressing the asymmetrical reciprocity between small proxy models and large client models. However, our model, \ours, differs from FedType in several key aspects.
First, the approach to calculating loss values is fundamentally different. FedType relies on training two conformal models to estimate uncertainty sets—one for the large client model and another for the proxy model. This process introduces additional computational overhead and requires tuning more hyperparameters. In contrast, \ours adopts a more straightforward strategy by directly leveraging logit ranks to select top classes, eliminating the need for extra conformal models.
Second, \ours employs a distinct loss function specifically designed for small clients, as described in Eq.~\eqref{eq:small_loss}, to address challenges such as small data sizes and low data quality. FedType, on the other hand, applies the same loss function uniformly across all clients, disregarding these specific challenges. This fundamental difference highlights the adaptability of \ours to heterogeneous client conditions.

\section{Methodology}

\subsection{Model Overview}
The goal of this work is to enable the training of federated learning under the settings of clients with different capacities. 
Let $\mathcal{C}^s = \{C^s_1, \cdots, C_{N_s}^s\}$ be the small client set in underserved regions, where $N_s$ represents the number of small clients.
Let $\mathcal{C}^l = \{C^l_1, \cdots, C_{N_l}^l\}$ denote the large client set, where $N_l$ denotes the number of large clients. 
Each client stores a training dataset $\mathcal{D}^s_i$ for a small client or $\mathcal{D}^l_j$ for a large client. In our setting, the size of $\mathcal{D}^l_j$ is far greater than that of $\mathcal{D}^s_i$. 
To increase the diagnostic ability of small clients, we propose a novel yet general framework \ours consisting of three key components: knowledge acquisition, small client training, large client training, and global model learning, in Figure~\ref{fig:overview}. 

The knowledge acquisition module aims to generate logits or probability distributions from $M$ foundation models $\{\mathbf{F}_1, \cdots, \mathbf{F}_M\}$ for public data $\mathcal{D}_p$, which are further used to guide the learning of clients. Note that the public data, including their image types and labels, may differ from those stored on clients.
For small clients, we design a new knowledge-guided surrogate training strategy to handle the issue of data insufficiency. Since the large clients hold high-quality and plenty of training data, we propose a novel asymmetrical dual knowledge distillation technique to distill large client models and lightweight proxy models. The lightweight models from both small and large clients will be uploaded to the server for global model learning. Next, we use the medical image classification task as an example and provide the details of each component.

\subsection{One-time Knowledge Acquisition from Foundation Models}\label{sec:knowledge_acq}
It is well-known that foundation models~\cite{dosovitskiy2020image,radford2021learning} usually outperform basic deep learning models on many tasks due to their large capacity. Unfortunately, more and more such models are packed as application programming interfaces (APIs) and not open-sourced like GPT-4. An ideal way to use these APIs is to directly upload a small set of data to their cloud servers, which helps us to fine-tune customized models and return them to users. However, in our setting, medical data are extremely sensitive and cannot be sent to third parties. Thus, it is challenging to obtain customized models without sharing private medical data.

To solve this challenge, we acquire knowledge from large foundation models with the help of public data $\mathcal{D}_p$, where all clients can access them. Assume that all the clients can also access the APIs of foundation models $\{\mathbf{F}_1, \cdots, \mathbf{F}_M\}$ and request the probability distributions for the public data. The returned probability distributions will be treated as knowledge for guiding the clients' training. 

\subsection{Small Clients: Knowledge-guided Surrogate Model Training}\label{sec:small_training}
In our setting, each small client can be treated as a rural emergency hospital holding a small set of data $\mathcal{D}^s_i$ without sufficient computational resources to train a sizeable yet accurate deep learning model. To solve this issue, we propose to use the knowledge returned from public APIs as guidance to help the model training. \emph{It should be emphasized that our model leverages foundation models exclusively through one-time API access during the initial phase of federated learning. This approach eliminates the need for deploying these models individually across each smaller client and does not involve them directly in the learning processes.}

Our general setting allows the accessible public data $\mathcal{D}_p$ to be different from the private data $\mathcal{D}_i^s$. Thus, the client model $\mathbf{w}^s_i$ cannot be directly used for training $\mathcal{D}_p$. To address this problem, we propose to introduce an auxiliary model $\hat{\mathbf{w}}^{s}_i$ for training on public data. $\hat{\mathbf{w}}^{s}_i$ shares the same feature extraction layers with the client model ${\mathbf{w}}^{s}_i$, which can be seen as a surrogate model of large foundation models. The only difference between $\hat{\mathbf{w}}^{s}_i$ and ${\mathbf{w}}^{s}_i$ is the classification layer that handles diverse data distribution. 

Specifically, \ours first queries an API $\mathbf{F}_m \in \{\mathbf{F}_1, \cdots, \mathbf{F}_M\}$ for each public data $\mathbf{x}_p \in \mathcal{D}_p$, which returns a label distribution $\mathbf{F}_m(\mathbf{x}_p)$. These returned distributions from $M$ APIs are used as guidance in surrogate model training via the following loss:
\begin{equation}\label{eq:small_pub_classification}\small
    \mathcal{R}^s_i = \sum_{p=1}^{P} [\text{CE}(\hat{\mathbf{w}}^{s}_i(\mathbf{x}_p), \mathbf{y}_p)
    +\lambda_R \text{KL}(\sum_{m=1}^M \alpha^m_{i,p}\mathbf{F}_m(\mathbf{x}_p)) || \hat{\mathbf{w}}^{s}_i(\mathbf{x}_p)],
\end{equation}
where $P$ is the number of public data, $\text{CE}(\cdot,\cdot)$ is the cross-entropy loss, and $\text{KL}(\cdot, \cdot)$ is the Kullback–Leibler divergence. $\lambda_R$ is the hyperparameter. $\alpha_{i,p}^m$ denotes the learned contribution score of each API on each public data, and $\sum_{m=1}^M \alpha^m_{i,p} =1$. 

\ours then trains the surrogate model ${\mathbf{w}}^{s}_i$ using the client private data $\mathcal{D}^s_i$ with the cross-entropy loss as follows:
\begin{equation}\label{eq:small_private_loss}
\mathcal{L}^s_i = \sum_{k=1}^{K_i^s} \text{CE}({\mathbf{w}}^{s}_i(\mathbf{x}^{s,i}_{k}), \mathbf{y}^{s,i}_{k}),
\end{equation}
where $K_i^s$ is the number of data in $\mathcal{D}^s_i$.
Since $\hat{\mathbf{w}}^{s}_i$ and ${\mathbf{w}}^{s}_i$ share the same feature extractor, we use a joint optimization approach by jointly optimizing $\mathcal{R}^s_i$ and $\mathcal{L}^s_i$ simultaneously via
\begin{equation}\label{eq:small_loss}
    \mathcal{J}_i^s = \mathcal{L}^s_i + \lambda_J  \mathcal{R}^s_i,
\end{equation}
where $\lambda_J$ is a trade-off parameter.

\begin{table*}[t!]
\centering
\caption{Accuracy comparison on the Fed-ISIC19 dataset. ``\underline{Underline}'' indicates the best baseline performance, and ``\textbf{bold}'' denotes the best performance. ``\% imp.'' is the value of percentage improvement compared with the best baseline.}
\label{tab:fed_isic19_result}
\begin{tabular}{c|l||c|c|c|c|c|c||c} 
\toprule 
\multirow{2}{*}{\textbf{Setting}}& \multirow{2}{*}{\textbf{Model}} & \multicolumn{3}{c|}{\textbf{Large}} & \multicolumn{3}{c||}{\textbf{Small}} & \multirow{2}{*}{\makecell[c]{\textbf{Client}\\\textbf{Average}}}\\\cline{3-8}
& & \textbf{Client 1} & \textbf{Client 2} & \textbf{Client 3} &\textbf{Client 4}&\textbf{Client 5}& \textbf{Client 6}&\\
\hline
\multirow{5}{*}{\rotatebox[origin=c]{90}{\textbf{\small{Homo.}}}}&FedAvg &0.4966		&0.4820	&0.4574	&0.3955 &0.3903	&0.2345	&0.4094\\
&FedProx  & 0.4895		&0.4871	&0.4612&0.4086 &	0.4015	&0.2411&	0.4148\\
 &Per-FedAvg & 0.5093	&0.4903	&0.4689	&	0.4126&0.4087&	0.2456&	0.4226\\
& PFedMe&0.5144		&0.5066&	0.4707	&0.4233&0.4153	&0.2508	&0.4302\\
 &PFedBayes&	\underline{0.5365}&	\underline{0.5207}&	\underline{0.4876}&	\underline{0.4278}	&0.4222	&0.2688&	\underline{0.4439}\\
 \hline
 \multirow{6}{*}{\rotatebox[origin=c]{90}{\textbf{\small{Hete.}}}}&FedMD &0.5248&	0.5122	&0.4603&	0.4139&	0.4005	&\underline{0.2718}&	0.4306\\
&FedGH  & 0.5286		&0.5014&	0.4755&0.4184	&0.4011&	0.2641	&0.4315\\
&FedKEAF & 0.5101		&0.4979&	0.4668&0.4096&	0.4121	&0.2669	&0.4272
\\
&FCCL  &  0.5175	&0.4961&	0.4614&	0.4107&	\underline{0.4248}	&0.2715	&0.4303
\\\cline{2-9}
&{\ours} &\textbf{0.5922} &\textbf{0.5563}	&\textbf{0.5322}&	\textbf{0.4956}&\textbf{0.4788}	&\textbf{0.3996}	&\textbf{0.5091}\\
&\small (\% imp.) & \small 10.38\%$\uparrow$ & \small 6.84\%$\uparrow$ & \small 9.15\%$\uparrow$ & \small 15.85\%$\uparrow$ & \small 12.71\%$\uparrow$ & \small 47.02\%$\uparrow$ & \small 14.69\%$\uparrow$ \\
\bottomrule 
\end{tabular}

\end{table*}

\subsection{Large Clients: Asymmetrical Dual Knowledge Distillation}\label{sec:large_training}
Different from small clients, large clients have sufficient data and computation resources to train complex deep learning models. Thus, it is unnecessary to utilize public data as small clients do, avoiding introducing noise during model training due to the different data distributions. 
However, uploading and downloading these large models consume a large amount of communication. To make the whole system more communication-efficient, we propose to distill small proxy models for these clients, which are further used in global model learning. 

Intuitively, the large capacity and strong predictive ability of the large model  $\mathbf{w}_j^l$ allows it to distill a small, powerful proxy model $\tilde{\mathbf{w}}_j^l$ with traditional knowledge distillation techniques. 
The large model $\mathbf{w}_j^l$ can be treated as a teacher, and the proxy model $\tilde{\mathbf{w}}_j^l$ can be seen as a student. However, such a simple approach aims to learn effective student models but ignores the importance of the proxy model.
In fact, the proxy model $\tilde{\mathbf{w}}_j^l$ contains two kinds of information. The first part is from the forward knowledge distillation, and the second is from other clients via the global model aggregation, which can be found in Section~\ref{sec:global_model_aggregation}. In other words, $\tilde{\mathbf{w}}_j^l$ carries diverse critical information aggregated from small clients $\mathcal{C}^s = \{C^s_1, \cdots, C_{N_s}^s\}$ and other large clients $\mathcal{C}^l_{\neq j} = \{C^l_1, \cdots, C_{j-1}^l, C_{j+1}^l, \cdots, C_{N_l}^l\}$. 

To tackle this issue, we design a novel asymmetrical dual knowledge distillation strategy, which enables the transfer of information in a bidirectional way -- \textit{forward} and \textit{backward}. 
The \textbf{forward} direction allows the information transfer from the large model $\mathbf{w}_j^l$ to the proxy model $\tilde{\mathbf{w}}_j^l$ with traditional knowledge distillation as follows:
\begin{equation}\label{eq:large_forward}
    \overrightarrow{\mathcal{KD}}_j^l = \sum_{k=1}^{K^l_j}\text{KL}(\mathbf{w}_j^l(\mathbf{x}^{l,j}_k) || \tilde{\mathbf{w}}_j^l(\mathbf{x}^{l,j}_k)),
\end{equation}
where $K^l_j$ is the number of data in $\mathcal{D}_j^l$. 
Since the large model $\mathbf{w}_j^l$ is usually more powerful than the proxy one $\tilde{\mathbf{w}}_j^l$, mandatorily distilling knowledge from $\tilde{\mathbf{w}}_j^l$ to $\mathbf{w}_j^l$ in the \textbf{backward} direction will introduce noise for the large model training. 
To address this problem, we propose a ranking-based knowledge distillation to imitate the \emph{behavior} of the proxy model for the large model. 

Intuitively, if the behaviors of the two models are similar, their prediction logits should also be similar. To avoid introducing extra noise by forcing the large model's logits to be similar to those of small ones, we propose to use the value rank of classes in the logits to imitate behaviors. The ranks only exhibit the relative magnitude of probabilities instead of real values, which can be treated as a loose constraint. 

Specifically, for a given data $\mathbf{x}_{k}^{l,j}$, the proxy model $\tilde{\mathbf{w}}_j^l$ can generate a logit or a class probability distribution $\tilde{\mathbf{w}}_j^l(\mathbf{x}_{k}^{l,j})$. Since our goal is to transfer the diversity information from the proxy model to the large one and avoid introducing too much extra information, in our proposed RKD, we only focus on the top-ranked classes in $\tilde{\mathbf{w}}_j^l(\mathbf{x}_{k}^{l,j})$. To make the large model imitate the behavior of the proxy model, we enforce to improve the ranks of these top classes in $\mathbf{w}_j^l(\mathbf{x}_{k}^{l,j})$ using the following loss:
\begin{align}
    \overleftarrow{\mathcal{KD}}_j^l = -\sum_{k=1}^{K_l^j} \sum_{r \in \Omega} \log(\frac{\exp(\mathbf{w}_j^l(\mathbf{x}_{k}^{l,j})[r])}{\Phi})\label{eq:ranking_KD}, \\\Phi = \sum_{u \in \Omega} \exp(\mathbf{w}_j^l(\mathbf{x}_{k}^{l,j})[u]) + \sum_{v \in \Omega^\prime} \exp(\mathbf{w}_j^l(\mathbf{x}_{k}^{l,j})[v]),
\end{align}
where $\Omega$ denotes the top-ranked class indexes, and $\Omega^\prime$ represents the remaining classes. Note that the only function of Eq.~\eqref{eq:ranking_KD} is to use the top class ranks $\Omega$ generated by the proxy model to guide the improvement of the corresponding class probabilities learned by the large model. This constraint relaxes the hard constraint of the traditional knowledge installation and enables the large model to imitate the behaviors of the proxy model.

\ours can also train the large model $\mathbf{w}_j^l(\mathbf{x}_{k}^{l,j})$ using the labeled dataset $\mathcal{D}_j^l$ with the cross-entropy loss $\mathcal{L}_j^l$, similar to Eq.~\eqref{eq:small_private_loss}. Finally, the loss function for training large clients is defined as follows:
\begin{equation}\label{eq:large_loss}
    \mathcal{G}_j^l = \mathcal{L}_j^l + \lambda_F \overrightarrow{\mathcal{KD}}_j^l + \lambda_B \overleftarrow{\mathcal{KD}}_j^l,
\end{equation}
where $\lambda_F$ and $\lambda_B$ are hyperparameters. In our setting, we only need the proxy models to have the same network structure, but the network structures of large client models can be different, which increases the generalization ability of the proposed framework.

\subsection{Global Model Learning }\label{sec:global_model_aggregation}
The surrogate models $\{\mathbf{w}_1^s, \cdots, \mathbf{w}_{N_s}^s\}$ from small clients and the proxy models $\{\hat{\mathbf{w}}_1^l, \cdots, \hat{\mathbf{w}}_{N_l}^l\}$ from large clients will be uploaded to the server to exchange parameter information. Since these models have identical network structures, we can use any existing model aggregation approaches, such as FedAvg~\cite{mcmahan2017communication}, to learn the global model. The global model will be distributed to each client for the iterative update until \ours converges.   
Note that the proposed \ours is a general framework and can also be used for other tasks, such as the medical image segmentation task, and the details can be found in \textbf{Appendix A}. Besides, the whole training procedure of the proposed \ours can be found in \textbf{Appendix B}.

\section{Medical Image Classification}\label{sec:image_classification}
In this section, we validate the proposed \ours on the three medical image classification tasks, including a \textbf{multiclass} melanoma classification and a \textbf{binary} pneumonia classification using chest x-rays.

\subsection{Experimental Settings}
\subsubsection{Baselines}
In our setting, clients are divided into large and small clients. Thus, this is a heterogeneous federated learning scenario. Small clients use ResNet20 as their model, while large clients use ResNet110. 
To fairly evaluate the performance of the proposed \ours, we employ two sets of baselines:
\begin{itemize}
    \item \textbf{Homogeneous} baselines are traditional federated learning models, including FedAvg~\cite{mcmahan2017communication}, FedProx\cite{li2020federated}, Per-FedAvg~\cite{fallah2020personalized}, PFedMe~\cite{t2020personalized}, and PFedBayes~\cite{zhang2022personalized}, which employ the small model ResNet20 as the client model. 
    \item \textbf{Heterogeneous} baselines include FedMD~\cite{li2019fedmd}, FedGH~\cite{yi2023fedgh}, FedKEAF~\cite{yu2022resource}, and FCCL~\cite{huang2022learn}. They use both ResNet110 for large clients and ResNet20 for small clients and use the public data $\mathcal{D}_p$ as a part of model input. 
\end{itemize}
Note that we do not list pFedHR~\cite{wang2023towards} as a baseline since it maintains a personalized model for each client on the server. However, \ours and baselines do not have such a constraint.
The details of each model and its implementation can be found in \textbf{Appendix C}.

\subsubsection{Implementation} For the medical image classification task, we employ two foundation models trained on CLIP~\cite{radford2021learning}, including ViT-L/14~\cite{dosovitskiy2020image} and RN50x16~\cite{radford2021learning}\footnote{We selected these two models to simulate foundation model APIs as they were pretrained on the CIFAR-100 dataset, which serves as the public data in the medical image classification experiment.}. The public dataset $\mathcal{D}_p$ is CIFAR-100~\cite{krizhevsky2009learning}, and the number of public data is 10,000. The proxy model is ResNet20. We set $\lambda_R = 0.1$, $\lambda_J = 0.2$, $\lambda_F = 1$, and $\lambda_B = 0.2$. We use accuracy as the evaluation metric. We set the size of top-ranked classes $\Omega$ in Eq.~\eqref{eq:ranking_KD} as 3 for multi-class classification and 1 for binary classification. With the early stopping mechanism, we set the maximum communication rounds to 100. Furthermore, our proposed model offers flexibility in the choice of public datasets. In our main results, we utilize CIFAR-100, a non-medical dataset. In Sec.~\ref{sec:public}, we present results from the medical public dataset NCT-CRC-HE-100K.

\subsection{Melanoma Classification}\label{sec:melanoma}
The Fed-ISIC19\footnote{The links of the datasets used in the experiments can be found in \textbf{Appendix D}.} dataset~\cite{ogier2022flamby,codella2018skin,tschandl2018ham10000}, consisting of 23,247 dermoscopy images, is used to classify eight different types of melanoma.
Using the data partition of FLamby~\cite{ogier2022flamby}, we divide the six clients into three large and three small ones, where the number of training/testing data is 9,930/2,483, 3,163/791, 2,690/673, 655/164, 351/88, and 180/45, respectively. The cross-silo setting requires all clients to be involved in training at each communication round.

\subsubsection{Performance Comparision}
Table~\ref{tab:fed_isic19_result} displays the experimental findings on the Fed-ISIC19 dataset. Notably, our proposed \ours outperforms all baseline models, particularly showcasing remarkable improvement on small clients. For the smallest client, the observed percentage improvement is substantial, reaching up to 47.1\%. Additionally, a consistent trend is observed across all approaches, indicating better performance on large clients compared to their smaller counterparts. This aligns with expectations, as larger clients inherently possess more data and even employ larger models, as seen in the heterogeneous baselines.

Although heterogeneous baselines leverage larger client models and incorporate additional public data, their performance remains comparable to most homogeneous models. This observation underscores that the aggregation approaches of heterogeneous models might not be well-suited for our setting, potentially due to small clients impeding the learning progress of large clients.

In contrast to all other approaches, our proposed \ours not only integrates foundation models to enhance the learning of small clients but also introduces a novel asymmetric dual knowledge distillation method to boost the learning of large clients. As a result, it achieves the highest performance. These results unequivocally demonstrate the effectiveness of the proposed \ours.


\begin{figure}[!t]
  \centering
  \includegraphics[width=0.4\textwidth]{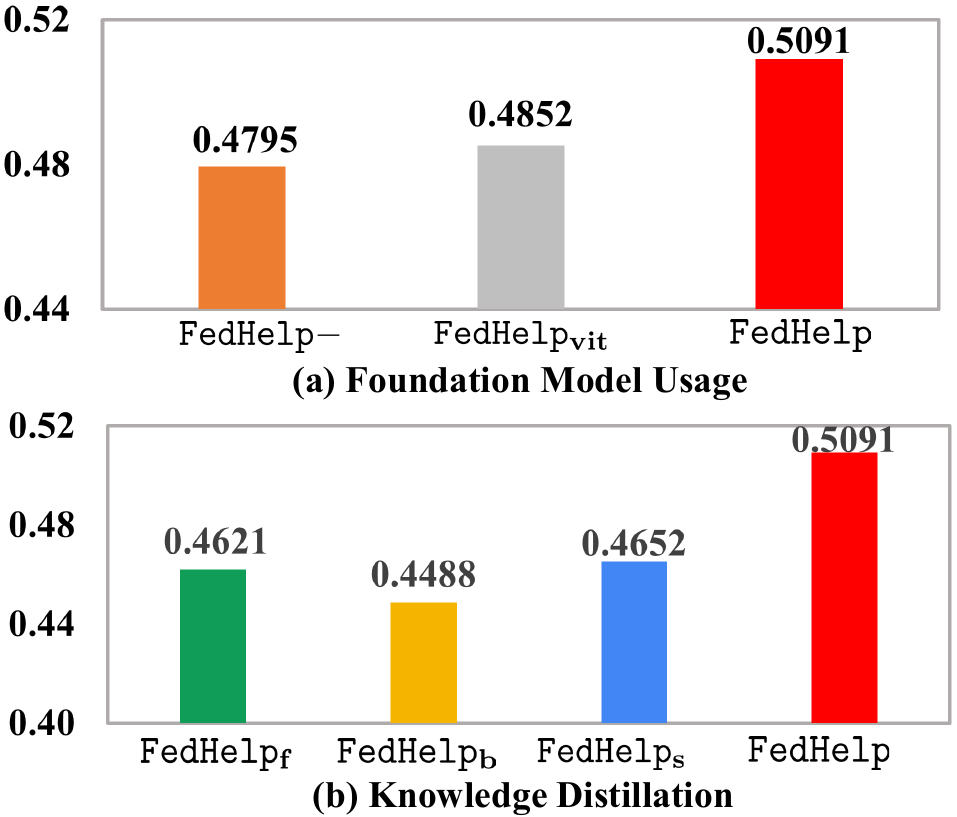}
  \caption{Averge client accuracy of two ablation studies on melanoma classification}
  \label{fig:classification_ablation}
\end{figure}

\subsubsection{Abalation Study}\label{sec:classification_ab}
We conduct two experiments to assess the effectiveness of our model design. 


(1) \textbf{Knowledge Enhancement with Foundation Models} (Section~\ref{sec:small_training}).
Given the integration of two foundation model APIs to enhance the learning of small clients, the first ablation study aims to validate the utility of these APIs. Two baselines are employed for comparison: The first baseline, denoted as \texttt{FedHelp}$-$, indicates the absence of any APIs, while the remaining components are identical to \ours.  \texttt{FedHelp}$_{vit}$ signifies the utilization of only one API, ViT, during the small client learning process. The results are shown in Figure~\ref{fig:classification_ablation} (a).
Our observations show that both {\texttt{FedHelp}$_{vit}$} and {\ours} outperform \texttt{FedHelp}$-$, indicating that the incorporation of foundation models is beneficial for enhancing the training of small clients. Furthermore, a positive correlation between the number of APIs and performance is evident when comparing \texttt{FedHelp}$_{vit}$ and \ours. Notably, even when removing all APIs (\texttt{FedHelp}$-$), the drop in performance is smaller than the increase observed in Table~\ref{tab:fed_isic19_result}. This result confirms that the primary performance improvement stems from the designed asymmetric dual knowledge distillation, rather than the use of foundation model APIs.

(2) \textbf{Asymmertic Dual Knowledge Distillation} (Section~\ref{sec:large_training}). In our model design, we propose a novel asymmetric dual knowledge distillation approach to benefit the learning of both small and large clients. We use three baselines in this ablation study to validate the effectiveness of the proposed strategy, including one-directional knowledge distillation methods (i.e., \texttt{FedHelp}$_{f}$ (forward) and \texttt{FedHelp}$_{b}$ (backward)) and a symmetric dual knowledge distillation approach \texttt{FedHelp}$_{s}$. The results are shown in Figure~\ref{fig:classification_ablation} (b). These results suggest that the three distillation approaches are not optimal for our setting. Solely employing forward knowledge distillation (\texttt{FedHelp}$_{f}$) effectively guides the learning of small models, but the transfer of knowledge from small to large models is lacking. Conversely, the outcome of \texttt{FedHelp}$_{b}$ indicates a potential lag in small model training compared to large model training, despite the inclusion of diverse information. Notably, even when using \texttt{FedHelp}$_{b}$, the foundation model APIs enhance our model's performance beyond the best baseline, as shown in Table~\ref{tab:fed_isic19_result}.
In contrast, traditional dual knowledge distillation (\texttt{FedHelp}$_{s}$) outperforms both \texttt{FedHelp}$_{f}$ and \texttt{FedHelp}$_{b}$, emphasizing the importance of exchanging knowledge between large and small clients. However, it highlights the need for a well-designed approach to model knowledge transfer from small to large clients.

\begin{table}[t!]
\caption{Average of 3 epoch training time (in seconds) on the Fed-ISIC19 dataset.}
\begin{tabular}{lc}
\toprule
\textbf{Setting}  & \textbf{Time(s)} \\
\midrule
Training small clients with \ours&275.67\\
Training small clients only using the CE loss&251.00\\\hline
Training large models with \ours  &511.67\\
Training large models only using the CE loss & 434.00\\
Training large models with the BKD loss &479.67\\
\bottomrule
\end{tabular}
\label{tab:computation_costs}
\end{table}

\begin{table*}[t!]
\centering
\caption{Performance comparison on the pneumonia classification task.}
\begin{tabular}{c|l||c|c|c|c|c|c||c}  
\toprule 
\multirow{2}{*}{\textbf{Setting}}& \multirow{2}{*}{\textbf{Model}} & \multicolumn{2}{c|}{\textbf{Large}} & \multicolumn{4}{c||}{\textbf{Small}} & \multirow{2}{*}{\makecell[c]{\textbf{Client}\\\textbf{Average}}}\\\cline{3-8}
& & \textbf{Client 1} & \textbf{Client 2} & \textbf{Client 3} &\textbf{Client 4}&\textbf{Client 5}& \textbf{Client 6}&\\
\hline
\multirow{5}{*}{\rotatebox[origin=c]{90}{\textbf{\small{Homo.}}}}&FedAvg &0.7862&	0.7687&	0.7464	&0.7225&	0.6685	&0.6487	&0.7235\\
&FedProx  &0.8311&	0.7996&	0.7802&	0.7648&	0.6844	&0.6605	&0.7534\\
 &Per-FedAvg &0.8451&	0.8221	&0.7878	&0.7705&	0.7002&	0.6754	&0.7669\\
& PFedMe&0.8523	&0.8183	&\underline{0.7932}&	0.7619&	\underline{0.7177}	&\underline{0.7163}&	0.7766
\\
 &PFedBayes&\underline{0.8569}	&\underline{0.8455}&	0.7869	&\underline{0.7852}&	0.7103	&0.7085	&\underline{0.7822}
\\
 \hline
 \multirow{6}{*}{\rotatebox[origin=c]{90}{\textbf{\small{Hete.}}}}&FedMD &0.8456&	0.8289&	0.7905&	0.7678&	0.6948&	0.6789	&0.7678\\
&FedGH  &0.8377	&0.8163&	0.7841	&0.7612	&0.6892&	0.6647&	0.7589\\
&FedKEAF & 0.8301	&0.8114&	0.7857	&0.7584	&0.6867	&0.6705&	0.7571\\
&FCCL  &0.8398&	0.8024	&0.7898	&0.7608&	0.6946	&0.6680&	0.7592
 \\\cline{2-9}
&\ours &\textbf{0.9044}&	\textbf{0.8846}&	\textbf{0.8455}	&\textbf{0.8032}&	\textbf{0.8011}&	\textbf{0.7645}	&\textbf{0.8338}
\\
&\small (\% imp.) & \small 5.54\%$\uparrow$ & \small 4.62\%$\uparrow$ & \small 6.60\%$\uparrow$ & \small 2.29\%$\uparrow$ & \small 11.62\%$\uparrow$ & \small 6.73\%$\uparrow$ & \small 6.60\%$\uparrow$ \\
\bottomrule 
\end{tabular}
%

\label{tab:pne}
\end{table*}

\subsubsection{Resource Usage Analysis}
Previous experiments have unequivocally illustrated the efficacy of our proposed approach, \ours. Nevertheless, leveraging the logits from public data for small client training and integrating asymmetrical reciprocity learning for large clients could further optimize resource utilization. This experiment aims to quantitatively assess resource consumption by scrutinizing both computation and communication costs.

(1) \textbf{Computation Costs.} 
Training time is a quantitative metric for assessing the efficiency of different approaches. In this experiment, we calculate the average training time across three epochs for various methods, with the results presented in Table~\ref{tab:computation_costs}. Here, ``BKD'' denotes the utilization of symmetric knowledge distillation during model training.
Analysis reveals that the training time for \ours surpasses that of baseline methods for both large and small clients. Nevertheless, the incremental training time is moderate, representing a 6.67\% increase compared to the large client using the BKD loss and a 9.83\% extension relative to the small client using the CE loss. Despite this, our approach demonstrates substantial performance enhancements, particularly for small clients, showcasing an improvement of up to 47.02\%, as illustrated in Table~\ref{tab:fed_isic19_result}. Importantly, these improvements persist even when employing only the logits from public data as guidance in small client learning.

(2) \textbf{Communication Costs.} In this medical classification task, substituting the large model (ResNet110 with 1.73 million parameters) with the proxy model (ResNet20 with 0.27 million parameters) results in an approximately 84.39\% reduction in communication costs per round.
In summary, while the proposed model does slightly increase computational costs, the reduction in communication costs and substantial performance improvements strongly underscore its notable advantages.

\begin{table}[t!]
\caption{Hyperprameter study on pneumonia classification task with chest x-ray dataset.}
\begin{tabular}{c|ccc}
\toprule
\textbf{Hyperparameter}  &\textbf{0.2}&\textbf{0.4}&\textbf{0.6} \\
\midrule
$\lambda_J$ &0.8338&0.8406&0.8389\\
$\lambda_B$&0.8338&0.8311&0.8276\\
\bottomrule
\end{tabular}
\label{tab:hyper_study}
\end{table}

\subsection{Pneumonia Classification}

\subsubsection{Performance Comparision}
We perform a binary classification evaluation for pneumonia using 5,863 chest x-ray images~\cite{wang2017chestx}. The dataset is divided among two large clients and four small clients, with the distribution of training and testing data as follows: 3,134/374, 1,048/124, 422/49, 317/37, 213/24, and 109/12, respectively.
The experimental results are presented in Table~\ref{tab:pne}. Similar patterns to those observed in Table~\ref{tab:fed_isic19_result} emerge, once again affirming the effectiveness of \ours.

\subsubsection{Hyperparameter Analysis}
We conduct a key hyperparameter study of our proposed approach, focusing on $\lambda_J$ for the surrogate model in Eq.~\eqref{eq:small_loss} and $\lambda_B$ for the large model in Eq.~\eqref{eq:large_loss}. We keep all other settings consistent with Table~\ref{tab:pne} in the paper. The average accuracy is shown in Table~\ref{tab:hyper_study}. As the value of $\lambda_J$ increases from 0.2 to 0.6, the performance initially improves and then declines. This trend may be due to the fact that appropriate utilization of public data enhances performance, but an over-reliance on it, at the expense of local data, can have a negative impact. Regarding $\lambda_B$, the performance declines slightly as it increases from 0.2 to 0.6. This could be because a larger $\lambda_B$ allows for more backward knowledge transfer from the surrogate model to the large model, which may be less quality than the knowledge passed from the large models to the surrogate models.

\begin{table*}[t!]
\centering
\caption{Accuracy comparison with the medical public dataset (NCT-CRC-HE-100K). 
}
\begin{tabular}{c|c|l||c|c|c|c|c|c||c} 
\toprule 
\multirow{2}{*}{\textbf{Task}}&
\multirow{2}{*}{\textbf{Setting}}& \multirow{2}{*}{\textbf{Model}} & \multicolumn{3}{c|}{\textbf{Large}} & \multicolumn{3}{c||}{\textbf{Small}} & \multirow{2}{*}{\makecell[c]{\textbf{Client}\\\textbf{Average}}}\\\cline{4-9}
& & & \textbf{Client 1} & \textbf{Client 2} & \textbf{Client 3} &\textbf{Client 4}&\textbf{Client 5}& \textbf{Client 6}&\\
\hline

\multirow{11}{*}{\rotatebox[origin=c]{90}{\textbf{Melanoma}}}
&\multirow{5}{*}{\rotatebox[origin=c]{90}{\textbf{\small{Homo.}}}}&FedAvg &0.4966		&0.4820	&0.4574	&0.3955 &0.3903	&0.2345	&0.4094\\
&&FedProx  & 0.4895		&0.4871	&0.4612&0.4086 &	0.4015	&0.2411&	0.4148\\
& &Per-FedAvg & 0.5093	&0.4903	&0.4689	&	0.4126&0.4087&	0.2456&	0.4226\\
&& PFedMe&0.5144		&0.5066&	0.4707	&0.4233&0.4153	&0.2508	&0.4302\\
& &PFedBayes&	0.5365&	0.5207&	\underline{0.4876}&	0.4278	&0.4222	&0.2688&0.4439\\
 \cline{2-10}
& \multirow{6}{*}{\rotatebox[origin=c]{90}{\textbf{\small{Hete.}}}}&FedMD &0.5377	&0.5287	&	0.4644		&0.4395	&0.4207&	\underline{0.2932}&	0.4474\\
&&FedGH  &\underline{0.5409}	&\underline{0.5396}	&	0.4756		&\underline{0.4486}&	\underline{0.4303}	&0.2864&	\underline{0.4536}\\
&&FedKEAF &0.5261&	0.5120&		0.4733	&	0.4367	&0.4289	&0.2803	&0.4429\\
&&FCCL  & 0.5283&	0.5197	&	0.4705&		0.4304&	0.4277	&0.2784&	0.4425\\\cline{3-10}

&&{\ours} &\textbf{0.5945}&\textbf{0.5644}&\textbf{0.5478}&\textbf{0.5375}&\textbf{0.4803}&\textbf{0.4132}&\textbf{0.5229}\\
&&\small (\% imp.) & \small 9.91\%$\uparrow$ & \small 4.60\%$\uparrow$ & \small 12.34\%$\uparrow$ & \small 19.82\%$\uparrow$ & \small11.62\% $\uparrow$ & \small 40.93\%$\uparrow$ & \small15.28\% $\uparrow$ \\
\bottomrule 
\toprule

\multirow{2}{*}{\textbf{Task}} &
\multirow{2}{*}{\textbf{Setting}}& \multirow{2}{*}{\textbf{Model}} & \multicolumn{2}{c|}{\textbf{Large}} & \multicolumn{4}{c||}{\textbf{Small}} & \multirow{2}{*}{\makecell[c]{\textbf{Client}\\\textbf{Average}}}\\\cline{4-9}
& & & \textbf{Client 1} & \textbf{Client 2} & \textbf{Client 3} &\textbf{Client 4}&\textbf{Client 5}& \textbf{Client 6}&\\
\hline
\multirow{11}{*}{\rotatebox[origin=c]{90}{\textbf{Pneumonia}}}
&\multirow{5}{*}{\rotatebox[origin=c]{90}{\textbf{\small{Homo.}}}}&FedAvg &0.7862&	0.7687&	0.7464	&0.7225&	0.6685	&0.6487	&0.7235\\
&&FedProx  &0.8311&	0.7996&	0.7802&	0.7648&	0.6844	&0.6605	&0.7534\\
& &Per-FedAvg &0.8451&	0.8221	&0.7878	&0.7705&	0.7002&	0.6754	&0.7669\\
&& PFedMe&0.8523	&0.8183	&0.7932&	0.7619&	\underline{0.7177}	&\underline{0.7163}&	0.7766
\\
& &PFedBayes&0.8569	&\underline{0.8455}&	0.7869	&\underline{0.7852}&	0.7103	&0.7085	&\underline{0.7822}
\\
 \cline{2-10}
& \multirow{6}{*}{\rotatebox[origin=c]{90}{\textbf{\small{Hete.}}}}&FedMD &\underline{0.8570}	&0.8311	&\underline{0.8026}&	0.7748	&0.7109&	0.6853&	0.7770
\\
&&FedGH  &0.8492	&0.8264&	0.7955&	0.7693&	0.7125&	0.6768	&0.7716\\
&&FedKEAF & 0.8389	&0.8189&	0.7893&	0.7652&	0.6804&	0.6714	&0.7607
\\
&&FCCL  &0.8466	&0.8175	&0.8012&	0.7707&	0.7057&	0.6891	&0.7718
\\\cline{3-10}
&&\ours &\textbf{0.9065}&\textbf{0.8881}&\textbf{0.8506}&\textbf{0.8017}&\textbf{0.8044}&\textbf{0.7667}&\textbf{0.8368}\\
&&\small (\% imp.) & \small 5.78\%$\uparrow$ & \small 5.04\%$\uparrow$ & \small 5.98\% $\uparrow$ & \small 2.10\% $\uparrow$ & \small12.08\% $\uparrow$ & \small 7.04\%$\uparrow$ & \small 6.98\%$\uparrow$ \\
\bottomrule 
\end{tabular}
\label{tab:fed_isic19_result_skin}
\end{table*}

\subsection{Public Dataset Selection}\label{sec:public}

We also experimented to assess the sensitivity of public dataset selection. Initially, we fine-tuned the two foundation models on the NCT-CRC-HE-100K dataset, utilizing the last 10,000 images as the public data. 
The outcomes of the Fed-ISIC19 and pneumonia classification tasks are detailed in Table~\ref{tab:fed_isic19_result_skin}. 

Compared to other heterogeneous baselines utilizing the same medical public data, our proposed model \ours demonstrates superior performance on each client and yields higher average results with the medical public data.
When compared with the results using the CIFAR-100 dataset as the public data (refer to Table~\ref{tab:fed_isic19_result} and Table~\ref{tab:pne}), the performance of \ours also experiences a marginal boost. This enhancement is attributed to the medical public data sharing more features similar to the local data, providing valuable knowledge that enhances the training of the small local client models.
For baselines, the performance of FedMD and FedGH exhibits significant improvement when using public medical data. This enhancement is attributed to the similarity between the public data and private data, contributing to better consensus and thereby boosting local training.

Consequently, our experimental results underscore that our proposed approach is not heavily reliant on choosing the public dataset. However, incorporating medical data does have a slight positive impact on performance, particularly when local clients are engaged in medical-related tasks.

\section{Medical Image Semantic Segmentation}
The proposed \ours is a general framework that can be used for both medical image classification and segmentation tasks. In this section, we use two medical image semantic segmentation tasks -- lung segmentation and brain mask segmentation -- on both 2D and 3D images to validate the utility of our framework.

\subsection{Datasets}
The 2D lung segmentation dataset contains 704 images, and we distribute them to two large and one small clients with the following number of training/test data: 285/31, 285/31, and 65/7, respectively. 
The 3D brain T1 magnetic resonance images (MRIs) dataset is extracted from the Information extraction from Images (IXI) database. We still follow the data partition of Fed-IXI used by FLamby~\cite{ogier2022flamby} and treat two clients as large and the third one as small. The number of training/testing images is 249/62, 145/36, and 59/15, respectively. 
The public dataset $\mathcal{D}_p$ used in this experiment is the dermoscopic lesion image dataset in the 2016 ISIC Challenge, consisting of 900 binary mask images.


\begin{table*}[t!]
\centering

\caption{The segmentation task results.}
\resizebox{1\textwidth}{!}{
\begin{tabular}{l||c|c|c|c|c|c|c|c||c|c|c|c|c|c|c|c} 
\toprule 
\textbf{Task} & \multicolumn{8}{c||}{\textbf{2D Lung Segmentation}} & \multicolumn{8}{c}{\textbf{3D Brain Mask Segmentation}}\\\hline
\textbf{Data} & \multicolumn{2}{c|}{\textbf{Client 1}} & \multicolumn{2}{c|}{\textbf{Client 2}} &\multicolumn{2}{c|}{\textbf{Client 3}} &\multicolumn{2}{c||}{\textbf{Client Avg}} & \multicolumn{2}{c|}{\textbf{Client 1}} & \multicolumn{2}{c|}{\textbf{Client 2}} &\multicolumn{2}{c|}{\textbf{Client 3}} &\multicolumn{2}{c}{\textbf{Client Avg}}\\\hline
\textbf{Metric}&Acc&DC&Acc&DC&Acc&DC&Acc&DC&Acc&DC&Acc&DC&Acc&DC&Acc&DC\\
\hline
FedAvg &0.8558	&0.8316&	0.8717&	0.8505&	0.8045&	0.7720&	0.8440	&0.8180 
&0.7769&	0.7520&	0.7651&	0.7317	&0.7086	&0.6869	&0.7502&	0.7235\\
FedProx & 0.9034	&0.8827	&0.9145	&0.9003	&0.8696&	0.8423	&0.8958&	0.8751
&0.7926&	0.7745	&0.7955&	0.7721	&0.7357	&0.7009	&0.7746&	0.7492\\
Per-FedAvg &0.9221&	0.9057&	0.9187&	0.8976&	0.8687&	0.8493&	0.9032&	0.8842
&0.8405&	0.7987	&\underline{0.8168}&	0.7865&	0.7386	&0.7086&	\underline{0.7986}	&0.7646\\
PFedMe & \underline{0.9277}	&\underline{0.9144}&	0.9305	&\underline{0.9211}&	0.8736	&0.8625&	0.9106&	\underline{0.8993}
&0.8333&	0.7932	&0.8107&	0.7743&	\underline{0.7499}	&\underline{0.7205}	&0.7979&	0.7627\\
PFedBayes & 0.9146	&0.8905	&0.9216	&0.9088	&0.8604	&\underline{0.8688}	&0.8989	&0.8894
&0.8475	&0.8104&	0.8155&	\underline{0.7927}&	0.7214&	0.7052	&0.7948&	\underline{0.7694}\\
FedSM&0.9384	&0.9076	&\underline{0.9351}	&0.9046&	\underline{0.8751}&	0.8663&	\underline{0.9162}&	0.8928
& \underline{0.8488}&	\underline{0.8196}&	0.8147&	0.7784&	0.7264	&0.6963	&0.7966&	0.7637
\\\hline
FedMD&0.9203&	0.9011	&0.9257&	0.9062&	0.8679&	0.8404	&0.9046	&0.8826
&0.8388&	0.8052&	0.8122&	0.7863&	0.7259&	0.7014	&0.7923	&0.7643\\
FedGH  &0.9088	&0.8848	&0.9139&	0.8968&	0.8581	&0.8315	&0.8936&	0.8710
&0.8371&	0.7989&	0.8049&	0.7730&	0.7375	&0.6937	&0.7932	&0.7552\\
FedKEAF &0.9105&	0.8966&	0.9216	&0.9077&	0.8525	&0.8362&	0.8949	&0.8802
&0.7902	&0.7651&	0.7845&	0.7562&	0.7223	&0.6890	&0.7657&	0.7368\\
FCCL  &0.9167	&0.8905	&0.9195	&0.8944&	0.8493&	0.8320	&0.8952&	0.8723
&0.8065	&0.7678&	0.7906&	0.7654&	0.7344&	0.7107&	0.7771&	0.7479\\\hline
\ours  &  \textbf{0.9655}	&\textbf{0.9378}	&\textbf{0.9574}&	\textbf{0.9297}&	\textbf{0.9258}	&\textbf{0.9026}	&\textbf{0.9496}&\textbf{0.9234}
 & \textbf{0.8554}	&\textbf{0.8248}&\textbf{0.8386}&	\textbf{0.8295}&	
\textbf{0.7577}&	\textbf{0.7389}&	\textbf{0.8172}&	
\textbf{0.7977}\\
\small (\% imp.) & \small 4.07\%$\uparrow$ & \small 2.56\%$\uparrow$ & \small 2.38\%$\uparrow$ & \small 0.93\%$\uparrow$ & \small 5.79\%$\uparrow$ & \small 3.89\%$\uparrow$ & \small 3.65\%$\uparrow$ 
& \small 2.68\%$\uparrow$ & \small 0.78\%$\uparrow$ & \small 0.63\%$\uparrow$ & \small 2.67\%$\uparrow$ & \small 4.64\%$\uparrow$ & \small 1.04\%$\uparrow$ & \small 2.55\%$\uparrow$ & \small 2.33\%$\uparrow$& \small 3.68\%$\uparrow$\\
\bottomrule 
\end{tabular}
}

\label{tab:seg}
%
\end{table*}

\begin{figure*}[h!]
\centering
\includegraphics[width=0.9\textwidth]{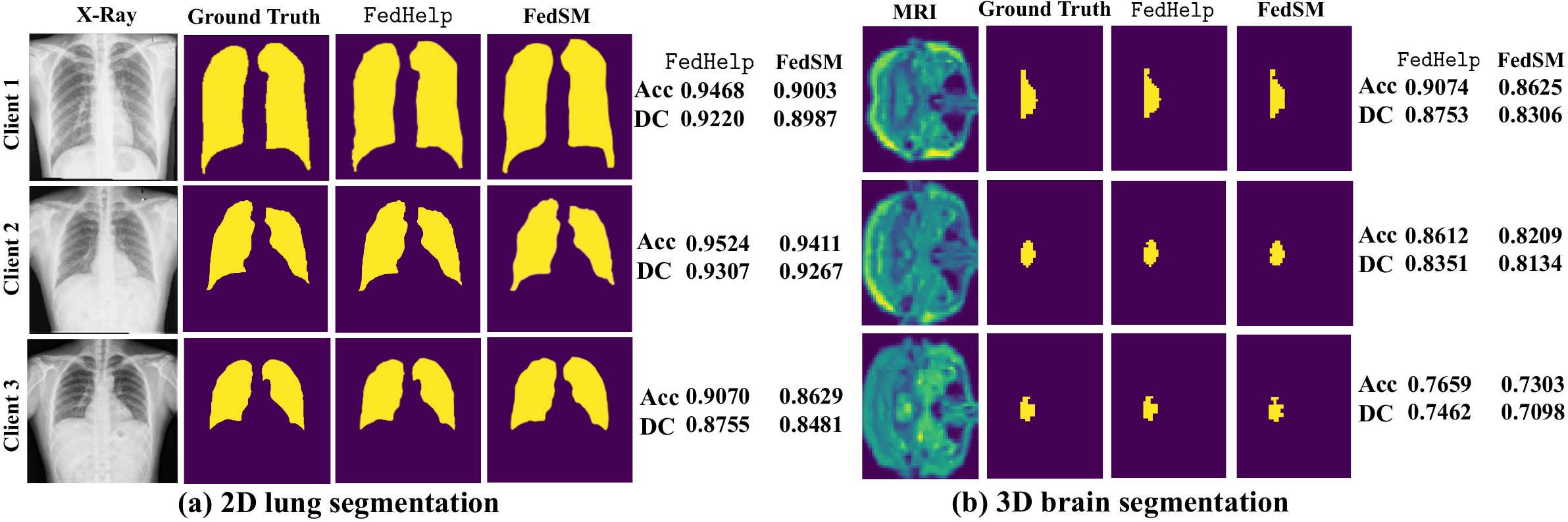}
    \caption{Visualization of 2D and 3D segmentation tasks.}
    \label{fig:seg_2d3d}
\end{figure*}

\begin{figure}[h]
  \centering
  \includegraphics[width=0.4\textwidth]{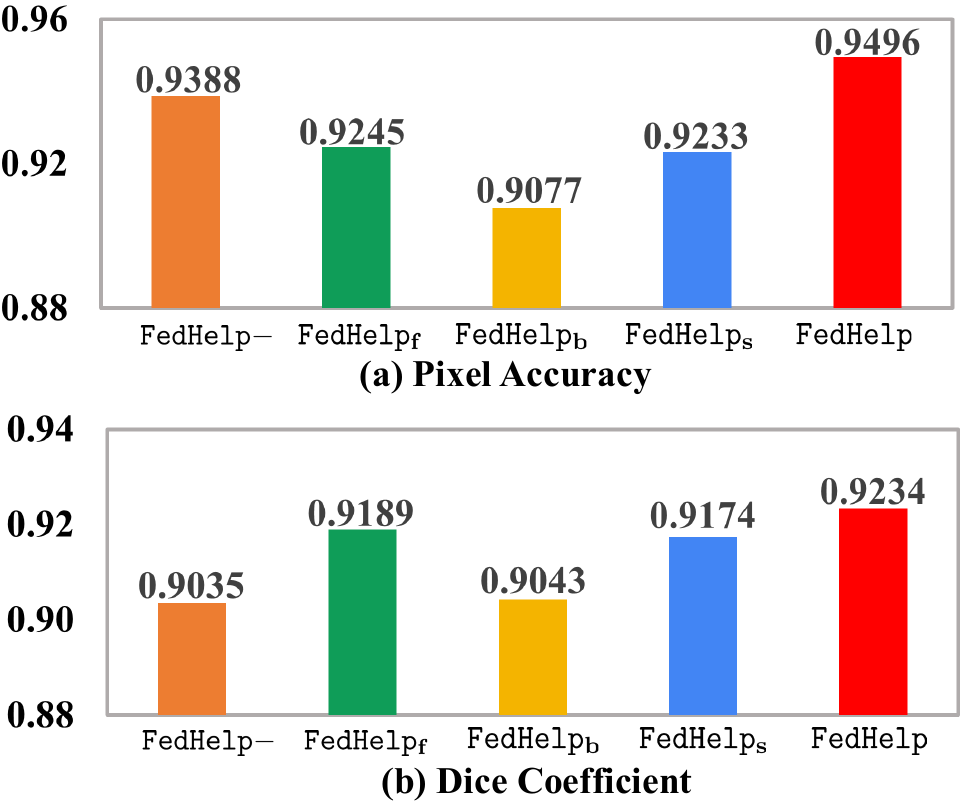}
  \caption{Abalation study results on segmentation task.}
  \label{fig:seg_ab}
\end{figure}

\subsection{Baselines and Implementations}
Except for the baselines used in Section~\ref{sec:image_classification}, we also add a federated learning-based medical image segmentation approach FedSM~\cite{xu2022closing} as a homogeneous baseline. 
We use MedSAM~\cite{MedSAM} as the foundation model API\footnote{\url{https://github.com/bowang-lab/MedSAM}} for segmentation tasks. 
We set $\lambda_R = 0.1$, $\lambda_J = 0.2$, $\lambda_F = 1$,  $\lambda_B = 0.2$, $\beta_0= 10$, and $\sigma = 5$. We use accuracy as the evaluation metric. We set the size of top-ranked classes $\Omega$ in Eq.~\eqref{eq:seg_backward} as 1 for the two segmentation tasks. The evaluation metrics are pixel accuracy and the Dice coefficient, following ~\cite{jiang2023fair,liu2021feddg,xu2022closing}. To ensure a fair comparison, an early stop mechanism is employed for each model training, and we report the average performance of the latest 10 models before reaching convergence for all approaches.

\subsection{Performance Analysis}
Table~\ref{tab:seg} presents the experimental results for two segmentation tasks. It is evident that the proposed \ours consistently outperforms all baseline models. Notably, FedSM~\cite{xu2022closing}, designed specifically for medical image segmentation tasks, demonstrates superior performance compared to our baselines. Similar to the medical classification task findings, homogeneous models tend to outperform heterogeneous ones in other baselines. We also randomly select one input image from each client and visualize the segmentation results in Figure~\ref{fig:seg_2d3d}. These results reaffirm the effectiveness and generalization ability of \ours.


\subsection{Ablation Study}
Similar to the medical classification, we also conduct ablation studies to validate the utility of each proposed module. Since only one API, MedSAM~\cite{MedSAM}, is used in this task, we keep \texttt{FedHelp}$-$ only to validate the influence of foundation models. \texttt{FedHelp}$_{f}$, \texttt{FedHelp}$_{b}$, and \texttt{FedHelp}$_{s}$ are kept to validate the proposed asymmetric dual knowledge distillation strategy. The observations of 2D lung segmentation (shown in Figure~\ref{fig:seg_ab}) are similar as discussed in Section~\ref{sec:classification_ab}.

\section{Conclusion}
This paper addresses the challenge of geographic health disparities in underserved regions with the aim of enhancing healthcare quality, by employing advanced federated learning techniques. We introduce a novel framework, named \ours, which harnesses the capabilities of foundation models to mitigate data insufficiency issues in underserved regions. Additionally, we propose a novel asymmetric dual knowledge distillation strategy to address the asymmetrical reciprocity among clients. Our experiments encompass both medical image classification (binary and multi-class labels) and segmentation tasks (2D and 3D). The experimental results confirm the effectiveness and utility of the proposed \ours, demonstrating its potential to ameliorate geographic health disparities. We are confident that this work will not only yield substantial benefits within the medical domain but will also deliver great value to businesses.

\section*{Acknowledgements}
The authors thank the anonymous referees for their valuable comments and helpful suggestions. This work is partially supported by the National Science Foundation under Grant No. 2238275 and 2348541.

\bibliographystyle{ACM-Reference-Format}
\balance
\bibliography{sample-base}

\section*{Appendix}

\subsection*{A. Medical Image Semantic Segmentation}

Here, we provide the details of \ours for the medical image segmentation task, which is similar to the medical image classification task but uses different loss functions for small and large client updates.

For the \textbf{small client training}, we still use APIs to obtain the pixel-level distributions $\{\mathbf{F}_m(\mathbf{x}_{p,q})\}_{q=1}^{Q_p}$, where $Q_p$ is the number of pixels in $\mathbf{x}_p$. After that, we can have the surrogate segmentation model training loss as follows:
\begin{align}\label{eq:small_pub_segmentation}
    \mathcal{R}^s_i = \sum_{p=1}^{P}\sum_{q=1}^{Q_p} \beta_{p,q}[\text{CE}(\hat{\mathbf{w}}^{s}_i(\mathbf{x}_{p,q}), \mathbf{y}_{p,q})
    \\+\lambda_R \text{KL}(\sum_{m=1}^M \alpha^m_{i,p}\mathbf{F}_m(\mathbf{x}_{p,q})) || \hat{\mathbf{w}}^{s}_i(\mathbf{x}_{p,q})],
\end{align}
where $\beta_{p,q}$ is a weight map to distinguish the importance of pixels in the training. Following U-Net~\cite{ronneberger2015u}, we pre-compute the weight map as follows:
\begin{equation}\label{eq:weigh_map_segmentation}
    \beta_{p,q} = \beta^c_{p,q} + \beta_o * \exp(-\frac{(d_1(\mathbf{x}_{p,q}) + d_2(\mathbf{x}_{p,q}))^2}{2\sigma^2}),
\end{equation}
where $\beta_{p,q}^c$ is the weight map to balance the class frequencies. $d_1$ and $d_2$ are the distances to the border of the nearest and the second nearest cell, respectively. $\beta_0$ and $\sigma$ are hyperparameters.
\ours then uses the private data to train the segmentation model using the loss as Eq.~\eqref{eq:small_loss}:
\begin{equation}\label{eq:small_seg_loss}
    \mathcal{L}^s_i = \sum_{k=1}^{K_i^s} \sum_{q=1}^{Q^{s,i}_k} \beta_{k,q}^{s,i}\text{CE}({\mathbf{w}}^{s}_i(\mathbf{x}^{s,i}_{k,q}), \mathbf{y}^{s,i}_{k,q}),
\end{equation}
where $\beta_{k,q}^{s,i}$ is the weight map that can be obtained with Eq.~\eqref{eq:weigh_map_segmentation}.

For the \textbf{large client training}, we still conduct the pixel-level classification via the proposed asymmetrical dual knowledge distillation. In the forward distillation, we still use traditional knowledge distillation for each pixel with the loss: 
\begin{equation}\label{eq:seg_forward}
    \overrightarrow{\mathcal{KD}}_j^l = \sum_{k=1}^{K^l_j}\sum_{q=1}^{Q^{l,j}_k}\text{KL}(\mathbf{w}_j^l(\mathbf{x}^{l,j}_{k,q}) || \tilde{\mathbf{w}}_j^l(\mathbf{x}^{l,j}_{k,q})).
\end{equation} 
The backward distillation is based on the top-ranked classes of each pixel as follows:
\begin{align}\label{eq:seg_backward}
\overleftarrow{\mathcal{KD}}_j^l = -\sum_{k=1}^{K_l^j} \sum_{q=1}^{Q^{l,j}_k}\sum_{r \in \Omega} \log(\frac{\exp(\mathbf{w}_j^l(\mathbf{x}_{k,q}^{l,j})[r])}{\Phi}),\\
\Phi = \sum_{u \in \Omega} \exp(\mathbf{w}_j^l(\mathbf{x}_{k,q}^{l,j})[u]) + \sum_{v \in \Omega^\prime} \exp(\mathbf{w}_j^l(\mathbf{x}_{k,q}^{l,j})[v]).   
\end{align}

Finally, following Eq.~\eqref{eq:large_loss}, we combine the cross-entropy loss $\mathcal{L}_j^l$ similar to Eq.~\eqref{eq:small_seg_loss}, the forward loss $\overrightarrow{\mathcal{KD}}_j^l$, and the backward loss $\overleftarrow{\mathcal{KD}}_j^l$, to train the segmentation models.

\subsection*{B. Algorithm Flow}

To provide a clear illustration of \ours, we outline the algorithmic flow in Algorithm~\ref{alg:alg_flow}. Notably, (1) we consolidate medical image classification and segmentation within Algorithm~\ref{alg:alg_flow}; (2) the small client update and large client update can be executed in parallel; and (3) obtaining logits of public data by querying the foundation models can be performed before model training, as they remain constant. Importantly, \ours optimizes communication costs by exclusively uploading and downloading small models, as indicated in lines 23 and 26.

\begin{algorithm}[ht]
\small
\SetAlgoLined
\DontPrintSemicolon
\Input{ Small and large client data $\{\mathcal{D}_1^s, \cdots, \mathcal{D}_{{N_s}}^s\}$ and $\{\mathcal{D}_1^l, \cdots, \mathcal{D}_{{N_l}}^l\}$, public data $\mathcal{D}_p$, communication rounds $T$, foundation models $\{\mathbf{F}_1, \cdots, \mathbf{F}_M\}$.} 
Initialize client models $\{\textbf{w}_{1,0}^s, \cdots, \mathbf{w}_{N_s, 0}^s\}$ and $\{\textbf{w}_{1,0}^l, \cdots, \mathbf{w}_{N_l, 0}^l\}$\\
Query $M$ APIs from foundation models $\{\mathbf{F}_1, \cdots, \mathbf{F}_M\}$ with public data $\mathcal{D}_p$;\\
\For{\textup{each communication round }$t = 1,2,\cdots$,T}{
    \kwClient{}{
    \# \textit{Small client update}\\
        \For{$i \in [1, \cdots, N_s]$}{
        \If{$t>1$}{$\mathbf{w}_{i,t}^s = \mathbf{w}^g_{t-1}$;}
            {Calculate $\mathcal{R}^s_{i,t}$ via Eq.~\eqref{eq:small_pub_classification}/Eq.~\eqref{eq:small_pub_segmentation} using the public data $\mathcal{D}_p$;\\
            Calculate $\mathcal{L}^s_{i,t} $ via Eq.~\eqref{eq:small_private_loss}/Eq.~\eqref{eq:small_seg_loss} using private data $\mathcal{D}_{i}$;\\
            \textup{Jointly optimize $\mathcal{R}^s_{i,t}$ and $\mathcal{L}^s_{i,t} $ by minimizing $\mathcal{J}_{i,t}^s$ via Eq.~\eqref{eq:small_loss};
            }
        }

            
        }
        \# \textit{Large client update}\\
        \For{$j \in [1, \cdots, N_l]$}{
        \If{$t>1$}{$\hat{\mathbf{w}}_{j,t}^l = \mathbf{w}^g_{t-1}$;}
            Calculate $\overrightarrow{\mathcal{KD}}_{j,t}^l$ via Eq.~\eqref{eq:large_forward}/Eq.~\eqref{eq:seg_forward};\\
            Calculate $\overleftarrow{\mathcal{KD}}_{j,t}^l$ via Eq.~\eqref{eq:ranking_KD}/Eq.~\eqref{eq:seg_backward};\\
            Update $\mathbf{w}^{l}_{j,t}$ by minimizing $\mathcal{G}_{j,t}^l$ via Eq.~\eqref{eq:large_loss};
        }

            


        
        Upload $\{\mathbf{w}_{1,t}^s, \cdots, \mathbf{w}_{N_s,t}^s\}$ and $\{\hat{\mathbf{w}}_{1,t}^l, \cdots, \hat{\mathbf{w}}_{N_l,t}^l\}$ to the server;
    
}

    \kwServer{}{
        \textup{Obtain the aggregated model $\mathbf{w}^g_t$ via FedAvg;}\\
        \textup{Distribute the aggregated model $\mathbf{w}^g_t$ back to clients.}
    }
    }

\caption{Algorithm Flow of \ours.}
\label{alg:alg_flow}
\end{algorithm}

\subsection*{C. Baselines}
In the \textbf{homogeneous} setting, we use the following approaches as the baselines:
\begin{itemize}
    \item FedAvg~\cite{mcmahan2017communication} trains a global model by averaging model updates from multiple decentralized clients, without sharing local data. Each client trains locally and sends updates to a central server, which averages these to improve the global model. The updated global model is then shared with all participants for further local training.
    \item FedProx~\cite{li2020federated} addresses the heterogeneity challenge in federated learning by generalizing and re-parameterizing FedAvg. 
    \item Per-FedAvg~\cite{fallah2020personalized} treats the global model as the initialization for the local client training and targets searching an optimal model initialization for each client model personalization within a few steps of local updates.
    \item PFedMe~\cite{t2020personalized} utilizes Moreau envelopes for client loss functions to guide the local model personalized learning, which decouples personalized model learning from global model learning.
    \item PFedBayes~\cite{zhang2022personalized} uses Bayesian variational inference to achieve personalized federated learning, which balances the data reconstruction error and KL divergence between local and global distributions during local updates.
    \item FedSM~\cite{xu2022closing} is designed for the medical segmentation task in federated learning, which addresses the generalization gap caused by client drift from non-iid data distribution.
\end{itemize}

\medskip
\noindent We use the following \textbf{heterogeneous} baselines:
\begin{itemize}
    \item FedMD~\cite{li2019fedmd} employs transfer learning and knowledge distillation, utilizing labeled public data on the server. Each client is required to train their local model using both public and private datasets. Subsequently, the clients transmit their class scores from the public dataset to the server, which then computes a consensus and sends it back to the clients for updating their models locally.
    \item FedGH~\cite{yi2023fedgh} allows clients to use individual feature extractors while sharing a uniform global header. Specifically, clients train their local models on personal data and send back both the representations and labels for each category to the server, facilitating the update of the global header. Following this, clients substitute their personal headers with the updated global header for making inferences.
    \item FCCL~\cite{huang2022learn} aims to tackle heterogeneity in federated learning by creating a cross-correlation matrix using unlabeled public data, aiding domain shift adaptation. It utilizes knowledge distillation in local updates to combat catastrophic forgetting while maintaining privacy. Clients send logits to the server for aggregation and then use the consensus logits to steer their local training processes.
    \item FedKEMF~\cite{yu2022resource} focuses on training diverse local models, facilitating effective knowledge integration, and implementing resource-conscious models. Clients transmit their network models to the server for a comprehensive knowledge distillation process. Subsequently, the server generates personalized models for each client and redistributes these for local updates. 
\end{itemize}

\subsection*{D. Dataset Repository}
We use the following datasets in our experiments:
\begin{itemize}
    \item Fed-ISIC19: \href{https://github.com/owkin/FLamby/tree/main/flamby/datasets/fed_isic2019}{\burl{https://github.com/owkin/FLamby/tree/main/flamby/datasets/fed\_isic2019}}.
    \item Pneumonia chest x-ray images: 
    \href{https://www.kaggle.com/datasets/paultimothymooney/chest-xray-pneumonia/data}{\burl{https://www.kaggle.com/datasets/\\paultimothymooney/chest-xray-pneumonia/data}}

    \item 2D lung segmentation dataset: 
    \href{https://www.kaggle.com/datasets/nikhilpandey360/chest-xray-masks-and-labels}{\burl{https://www.kaggle.com/datasets/\\nikhilpandey360/chest-xray-masks-and-labels/}}
    
    \item Information extraction from Images (IXI) database: \\\href{https://brain-development.org/ixi-dataset}{\burl{https://brain-development.org/ixi-dataset}}

    \item 3D Fed-IXI: 
    
    \href{https://github.com/owkin/FLamby/tree/main/flamby/datasets/fed_ixi}{\burl{https://github.com/owkin/FLamby/tree/main/flamby/datasets/\\fed\_ixi}}

    \item Dermoscopic lesion image dataset:
    
    \href{https://isic-challenge-data.s3.amazonaws.com/2016/ISBI2016_ISIC_Part1_Training_Data.zip}{\burl{https://isic-challenge-data.s3.amazonaws.com/2016/\\ISBI2016\_ISIC\_Part1\_Training\_Data.zip}}
    
    \item NCT-CR-HE-100K: 
    \href{https://paperswithcode.com/dataset/nct-crc-he-100k}{\burl{https://paperswithcode.com/dataset/\\nct-crc-he-100k}}
    
\end{itemize}

\end{document}